# Context information can be more important than reasoning for time series forecasting with a large language model

Janghoon Yang
*Computer Science Program.*
*Penn State Abington*
Abington, PA, USA
jxy5427@psu.edu

*Abstract*— With the evolution of large language models (LLMs), there is growing interest in leveraging LLMs for time series tasks. In this paper, we explore the characteristics of LLMs for time series forecasting by considering various existing and proposed prompting techniques. Forecasting for both short and long time series was evaluated. Our findings indicate that no single prompting method is universally applicable. It was also observed that simply providing proper context information related to the time series, without additional reasoning prompts, can achieve performance comparable to the best-performing prompt for each case. From this observation, it is expected that providing proper context information can be more crucial than a prompt for specific reasoning in time series forecasting. Several weaknesses in prompting for time series forecasting were also identified. First, LLMs often fail to follow the procedures described by the prompt. Second, when reasoning steps involve simple algebraic calculations with several operands, LLMs often fail to calculate accurately. Third, LLMs sometimes misunderstand the semantics of prompts, resulting in incomplete responses.

I. INTRODUCTION

With the introduction of the transformer module into deep learning, models for natural language processing have evolved rapidly. Initially, this module was used to build pretrained language models (PLMs) such as BERT [1] and XLNet [2]. Through transfer learning, after training the model with a large amount of data, PLMs have been fine-tuned for various tasks such as sentiment analysis, entity recognition, summarization, and translation. While fine-tuned PLMs often provide state-of-the-art (SOTA) performance for specific tasks and domains, their applicability to different types of tasks and data domains is quite limited. To overcome these limitations, large language models (LLMs) have been developed by leveraging large clusters of graphical processing units (GPUs), incorporating massive parameters and extensive training datasets. Although LLMs may perform inferior to SOTA models for some specific tasks in certain domains, they can handle a variety of tasks without further training and perform complex tasks

However, LLMs may have inherent limitation when the LLMs work with numerical problems without fine-tuning. The token generation in LLMs depends on the input and preceding output. This implies that when an algebraic operation is required, non-numeric tokens act as noise to the algebraic operation, hindering the model's ability to focus solely on the numerical aspects. In addition to this, there are several other issues with LLMs when numeric values in a query hold significant information for generating an accurate answer. This is particularly problematic when LLMs process tasks related to time series of numeric values, such as forecasting and outlier detection [3]. Inputs to LLMs are first tokenized, and many tokenizers build on byte pair encoding (BPE), which progressively expands the dictionary with frequently used subwords. Thus, numbers with long digits can be split into multiple tokens, hindering the extraction of precise patterns in numerical sequences and numerical operations. Furthermore, LLMs are trained to handle arbitrary strings with a massive corpus, which may limit their capability to handle complex numerical operations accurately unless they are fine-tuned with proper datasets. This is because they are not trained enough to identify specific patterns and rules in numerical data. It is noted that while numerical data appears to be a simple data type based on 10 digits, it can be of arbitrary length, and its meaning can vary greatly depending on the context in which the numeric values are generated. This irregularity and the large effective vocabulary size can pose additional challenges for LLMs in accurately executing tasks associated with time series.

It is also noted that the quality of generated output from LLMs often depends on the formulation of the questions, known as prompt design [4]. It can be expected that the quality of time series tasks with LLMs also depends on the prompt. However, unlike conventional natural language processing, time series tasks with LLMs can be more challenging as they require the implicit retrieval of procedures for a given task and accurate numeric algebraic operations to generate precise answers. In this regard, this paper explores the potential of prompts in time series forecasting with a LLM, GPT-4o-mini to elucidate the effect of prompts on time series tasks and shed light on the direction for exploiting LLMs for these tasks. To the best of the authors' knowledge, the behavior of LLMs with prompts for time series forecasting has not been thoroughly assessed both quantitatively and qualitatively, this study aims to fill the gap. To this end, we evaluated several existing general-purpose prompts and hand-crafted prompt based on seasonal ARIMA, statistical model for time series forecasting which we call SARIMA prompting. It was found that simply providing proper context information related to the time series, without additional reasoning prompts, can achieve performance comparable to the best-performing prompt for each case while one-shot chain of thought (CoT) prompting [5] and long short term (LST) prompting [6] achieves the best performance with some specific dataset. From analyzing the generated outputs qualitatively, several interesting characteristics in forecasting with LLMs were identified. LLMs often fail to follow the procedures described by the prompt. When reasoning steps involve simple algebraic calculations with numerous operations, LLMs often fail to calculate accurately. LLMs sometimes misunderstand the semantics of prompts, resulting in incomplete responses. For example, despite prompts requiring the use of trends over the entire time series, LLMs frequently fail to utilize trend information, identify seasonality, and exploit this information effectively. .

## II. RELATED WORKS

### A. Prompting

Many LLMs are trained to generate the next token based on the preceding sequence of tokens. Due to this intrinsic nature, the quality of the output from an LLM depends significantly on the design of the question, often leading to structured generation. Depending on the degree of additional programming required, which increases with complexity and interaction level, prompting can be classified into simple passive prompting, iterative prompting, and interactive prompting.

As a pioneering work in prompting for LLMs, chain of thought (CoT) prompting [5] was proposed to initiate reasoning in LLMs. This method often significantly improves the quality of the output by adding a simple phrase, 'let's think step by step.' Its simplicity and universal applicability, regardless of domain and tasks, have brought significant attention to prompting. To further improve, plan and solve (PS) prompting [7], which involves planning with subtasks and working on each subtask sequentially, has been shown to outperform CoT prompting in some tasks. It is noted that these simple passive promptings are constructed by adding a simple sentence to elicit reasoning for solving the given task.

These simple prompts can be further improved by adding a few demonstrations, a technique often referred to as few-shot learning. However, the effectiveness of these demonstrations depends on their level of articulation. To overcome this limitation, a systematic approach to designing demonstrations from LLM-generated content, involving multiple sequential prompts, can be employed. We call this method iterative prompting. Synthetic prompting [8] iteratively synthesizes queries from several initial demonstrations and generates responses with these synthesized queries, which are later used as demonstrations for chain-of-thought (CoT) prompting. This approach can be advantageous because internally generated queries and answers are more likely to align with the LLM's reasoning structure.

This iterative process can be further refined by introducing another model trained to generate demonstrations, which we call interactive prompting. Interacting models can be the same model with different roles [9] or different models with different capabilities [10]. This approach aims to improve prompts through interaction between multiple models, reducing human error in prompt articulation and bridging the cognitive gap between humans and LLMs in understanding information. Prompt with actor-critic editing (PACE) [10] exploits actor-critic structures popular in reinforcement learning. The actor model generates a prompt, while the critic model provides feedback on the generated prompt. While this method can improve the quality of the prompt for a specific task, the generated prompt's applicability becomes limited due to the nature of the generation. This interactive prompting has been utilized to generate prompts automatically, reducing human effort in creating prompts.

### B. Time Series Analysis with LLM

LLMs have very powerful generalizability, making them applicable to many different types of tasks. In line with this, time series can also be analyzed by LLMs. However, naïve application to time series analysis often results in performance inferior to SOTA performance. Thus, several different approaches to enhance the LLM's capabilities in analyzing time series have been made, considering the complexity. These approaches can be broadly classified into prompting, fine-tuning, and patching.

Prompting is a powerful technique to elicit a systematic procedure for completing a task. Therefore, it is very natural to consider prompting as a means to enhance the LLM's problem-solving capabilities in time series analysis. GPT-4 was shown to achieve better binary forecasting over stock data consisting of numerical and textual data by exploiting diverse data handling capabilities from manually articulated prompts than gradient boosting machine learning algorithms [11]. A universally applicable prompt for time series forecasting was proposed. LSTPrompt [6] was designed to integrate both short-term and long-term forecasting to help LLMs follow step-by-step reasoning more effectively.

The performance achieved by prompting can be limited due to the intrinsic nature of LLMs to textual information rather than numerical time series. Thus, a natural way to overcome this limitation is to fine-tune the LLM such that it can learn how to analyze the time series. PromptCast [12] decorates a time series with a template containing simple contextual information and metadata on the time series to create a prompt and fine-tunes the LLM in a question-answering way so that the LLM can learn how to predict for a given time series. To deal with the tokenization issue in numerical information, TIMELLM [13] fine-tunes the LLM by converting numbers into space-separated digits and scaling, achieving comparable performance to transformer-based models specifically designed for time-series tasks.

While fine-tuning makes LLMs better fitted for analyzing time series, the conventional tokenization based on conventional vocabulary constructed from textual corpus may not be good enough to represent time series efficiently. In this regard, state-of-the-art methods with LLMs for analyzing time series exploit patching by segmenting the time series into smaller chunks. Thus, to exploit LLMs' powerful inference capability and overcome the limitation in manipulating the time series as a sequence of numeric values, time series are patched first, embedded, and then aligned with the embedding space of the LLM. LLM-time [14] represents input time series with a prototype as input to LLM while numerical time series are patched and embedded by a separate network. While fixing the LLM, the combined network is trained to be fitted to time series tasks through aligning textual data and numeric data. TEMPO [15] created an input from decomposing time series into trend, seasonal, and residual components, patching each component separately, and concatenating with a soft prompt translated with learnable linear projection. Then, GPT-2 is fine-tuned over the position embedding layer, layer normalization layers, and attention layer with low-rank adaptation (LORA). This transforms the LLM into a model specialized for time series analysis.

Some other interesting approaches exploiting knowledge graphs and self-supervised learning were also proposed. An LLM can be indirectly utilized to construct a graph neural network [16] rather than analyzing time series directly. Specifically, an LLM can generate a training dataset to fine-tune another LLM, enabling it to produce quadruples that include temporal information for knowledge graph construction. This graph neural network then learns to predict the state of an entity at a given time by analyzing the patterns in the trajectories of entities and their relationships. Multi-modal contrastive learning, utilizing both textual descriptions and time series data, was also employed to cluster the data. This approach enables similarity searches to classify a given time series without the need for specific labels [17].

### III. PROMPTING

Prompting is a mechanism to formulate queries in a way that LLMs can respond better. While domain-specific and task-specific prompts often lead to desirable responses, crafting a proper prompt can be time-consuming. Recently, researchers have made efforts to devise universally applicable prompting. In line with this, a new prompting method for time series forecasting based on a statistical model, the SARIMA model, is proposed.

#### A. Baseline

A simple query structure for time series can be given as "{time_series} forecast next k steps," where {time_series} is a placeholder for a sequence of numeric values. However, LLMs can exploit their world knowledge, such as dependencies of physical phenomena on time. In this perspective, it is beneficial to add essential context information. To this end, we use the prompt format from PromptCast [12] as the base query format, considering the forecasting performance of various fine-tuned LLMs. This type of prompt includes context information such as domain and time resolution. The prediction task is also framed naturally in line with the domain. An illustrative example query is: " From April 15, 2020, Wednesday to April 29, 2020, Wednesday, the average temperature of region 110 was 44, 51, 59, 52, 51, 58, 64, 58, 60, 63, 60, 54, 53, 63, 66 degree on each day. What is the temperature going to be on April 30, 2020, Thursday?"

#### B. SARIMA Prompting

One of the major challenges in forecasting is non-stationarity. The Seasonal ARIMA model is known to be an efficient way to address this by considering trend and seasonality together [12]. TEMPO [15], which exploits time series decomposition and transformer modules, has been shown to achieve SOTA performance over many time-series datasets. Considering these recent efforts to add some inductive bias to the model, balancing human knowledge of time series characteristics and the powerful inference capabilities of LLMs, we propose a prompting method that imposes decomposition of time series forecasting, which we call SARIMA prompting. The underlying motivation of SARIMA prompting is that LLMs are good at solving problems when presented as a series of easier subproblems. Additionally, mandating LLMs to work on predicting each component separately is expected to improve forecasting in the presence of trends and seasonality. The initial draft for prompting was designed to consist of steps for decomposing the time series into trend, seasonality, and short-term variations, predicting each component, and merging the results to create a final prediction. This draft was then refined through several iterative updates with ChatGPT. The corresponding prompt is shown in Figure 1.

#### C. Prompting for comparison

Existing promptings were also considered for comparison. While CoT [5] and Plan and Solve [7] were proposed for general purposes, LSTPrompt [6] was designed specifically for time series forecasting.

**CoT** [5]: CoT prompting is a pioneering work that highlights the importance of prompting. Simply adding 'Think step by step' at the end of the query often elicits a reasoning procedure learned through training. However, calculation errors, missing-step errors, and misunderstanding errors were identified as major pitfalls with this prompting [7].

**PaS+** [7]: PaS was proposed to overcome missing-step errors. This prompting consists of a plan phase and a solve phase, with the expectation that the LLM will decompose a problem into subproblems during the plan phase and solve the problem step by step. PaS can be considered a refined CoT with additional instructions, 'Plan and Solve.' The PaS+, a detailed version of PaS, includes 'and show the **Final Answer** with predicted value only' to easily parse the predicted value'. The corresponding prompt is: 'A: Let's first understand the problem, extract relevant variables and their corresponding numerals, and make a plan. Then, let's carry out the plan, calculate intermediate variables (pay attention to correct numerical calculation and commonsense), solve the problem step by step, and show the **Final Answer** with predicted value only.'

**LST** [6]: While CoT and PaS+ can be applied to arbitrary tasks, the LST prompt was designed for time series forecasting. This prompt asks the LLM to solve the problem by integrating short-term and long-term forecasting. Additionally, it includes TimeBreath to reduce errors from cognitive load by segmenting the prediction horizon. It is closely related to the PaS+ prompt in that it exploits specific planning and solving steps tailored for time series forecasting. The prompts in the GitHub repository in [6] were used exactly the same way for fair comparison.

---

A: Let's analyze the sequence of numbers systematically to predict the next value. We will:
1. Decompose the sequence into three components:
• Trend: Identify the overall direction (increasing, decreasing, or stable).
• Seasonality: Detect repeating patterns or periodic fluctuations.
• Short-term variations: Examine irregularities or residual noise.
2. Predict each component by examining statistical values, such as mean, variance, and recent trends, ensuring accurate numerical calculations.
3. Combine the predicted components to form the final value.
For each step, calculate intermediate variables, justify the decisions using patterns and common sense, provide the reasoning clearly and step by step, and show the **Final Answer** .

---

Fig. 1. SARIMA Prompt

### IV. EXPERIMENTS

**Task**: A single next value prediction for a short time series and a next 6 values prediction for a long time series will be considered.

**Datasets**: The PISA dataset [15] was used for forecasting with a short time series, while the Individual Household Electric Power Consumption (IHEPC) dataset from the UCI Machine Learning Repository was used for forecasting with a long time series. The PISA dataset consists of three different time series: SG for the number of visitors over 15 days, CT for city temperature over 15 days, and ECL for electricity consumption over 15 days. The original IHEPC dataset was constructed from measurements taken per minute with 9 attributes. Due to the large variation per minute and the prediction horizon considered in this experiment, a single attribute, "Global_intensity" (average current per minute), was chosen and averaged over 60 minutes, resulting in 24 data points representing average hourly current usage per day. Subsequently, 3000 time series with a length of 96 data points were constructed by shifting every 10 data points. This dataset is expected to include seasonality and slight trends for a subset of series, and random variation over time associated with human activity.

**Tokenizer and Model**: The gpt-4o-mini-2024-07-18 by OpenAI, a distilled version of GPT-4o, was used as the base model for experimenting with different prompts, considering both cost and performance. The maximum output token size was set to 1024 for all prompts except LST prompting, for which it was 1280. This model uses the Tiktoken tokenizer, which is based on byte pair encoding (BPE). The tokenizer does not treat numeric values as a single token but includes vocabularies for numbers from 0 to 999, highlighting a fundamental limitation in LLMs for time series analysis. For example, the number 13245 is tokenized into two tokens: 132 and 45, while 12.992 is tokenized into three tokens: 12, ., and 992.

**Prompting**: The final query to the LLM was created by concatenating "Q:", the baseline prompt, and a prompt designed for each prompting method. Additionally, prompting with a one-shot example was also considered for CoT and SARIMA. The one-shot example was constructed with the following procedures: We manually checked 5 responses from each PISA dataset. Then, the one considered to be the best was selected by the author. It was refined to be applicable to time series independently of domain specificity by removing domain-specific terminologies such as temperature or power. This manually crafted example was further improved iteratively by asking OpenAI ChatGPT to generate a refined version, with the expectation that it may produce a prompt that ChatGPT can understand better. The one-shot example for SARIMA was designed similarly, except that a manually synthesized time series was used for numerical values so that the one-shot example could have an accurate answer for prediction at each component. For one-shot CoT and one-shot SARIMA, the final query was constructed by adding a zero-shot query at the end of the example. Corresponding examples can be available online [18].

**Performances**: The efficacy of prompting in forecasting was evaluated using root mean squared error (RMSE) and mean absolute error (MAE). Table I shows the performance with the SG dataset. Baseline prompting is observed to achieve the best performance. It is expected that the structure of the query was well designed for this dataset, even though it was originally designed to fine-tune the LLM. One exemplary query with baseline prompting is: "From August 21, 2021, Saturday to September 04, 2021, Saturday, there were 19, 17, 20, 17, 23, 14, 13, 18, 20, 14, 10, 17, 16, 18, 5 people visiting POI 324 on each day. How many people will visit POI 324 on September 05, 2021, Sunday?" The context of this query is that the numbers represent the daily count of people visiting a place. Additionally, there is date information that can potentially provide cues for forecasting. However, this sequence is unlikely to be influenced by any physical laws.

From this perspective, the superior performance with baseline prompting can be attributed to a simple thinking process that prevents large errors from overthinking. The large error with one-shot SARIMA can be ascribed to the inaccurate decomposition of each component. Manually checking the cases with large errors reveals that forecasting the seasonality component and short-term variation was often done without subtracting the trend or seasonality components from the time series, resulting in 2 to 3 times larger predicted values after combining the three component predictions. A negative value is forecasted by LST prompting, which occurs when the recent 2 or 3 values in time series are decreasing.

TABLE I. PERFORMANCE WITH SG DATASET (RMSE* AND MAE* ARE METRICS EVALUATED OVER THE SUBSET OF COMMON DATA)

| SG | RMSE | MAE | RMSE* | MAE* |
|---|---|---|---|---|
| Baseline | **9.665** | **6.634** | **9.845** | **6.720** |
| Zero-shot CoT | 10.341 | 7.232 | 10.565 | 7.298 |
| One-shot CoT | 10.766 | 7.622 | 10.955 | 7.679 |
| Zero-shot PaS+ | 10.314 | 7.172 | 10.529 | 7.252 |
| Zero-shot SARIMA | 10.797 | 7.383 | 11.226 | 7.524 |
| One-shot SARIMA | 17.006 | 10.173 | 17.443 | 10.259 |
| Zero-shot LST | 10.803 | 7.326 | 11.039 | 7.387 |

TABLE II. PERFORMANCE WITH CT DATASET (RMSE* AND MAE* ARE METRICS EVALUATED OVER THE SUBSET OF COMMON DATA)

| CT | RMSE | MAE | RMSE* | MAE* |
|---|---|---|---|---|
| Baseline | 8.050 | 6.148 | 8.149 | 6.237 |
| Zero-shot CoT | 8.469 | 6.283 | 15.180 | 6.488 |
| One-shot CoT | **7.355** | **5.369** | **7.847** | **5.500** |
| Zero-shot PaS+ | 9.461 | 6.451 | 9.357 | 6.536 |
| Zero-shot SARIMA | 8.588 | 5.836 | 8.758 | 5.942 |
| One-shot SARIMA | 17.828 | 10.201 | 17.717 | 10.208 |
| Zero-shot LST | 7.870 | 5.925 | 7.878 | 5.933 |

TABLE III. PERFORMANCE WITH ECL DATASET (RMSE* AND MAE* ARE METRICS EVALUATED OVER THE SUBSET OF COMMON DATA)

| ECL | RMSE | MAE | RMSE* | MAE* |
|---|---|---|---|---|
| Baseline | 649.746 | 412.468 | 677.533 | 428.214 |
| Zero-shot CoT | 717.085 | 453.458 | ####### | 470.021 |
| One-shot CoT | **611.890** | **383.457** | **647.198** | **401.620** |
| Zero-shot PaS+ | 678.373 | 441.652 | 712.805 | 458.866 |
| Zero-shot SARIMA | 652.708 | 410.029 | 679.480 | 421.748 |
| One-shot SARIMA | 3640.895 | 1224.527 | 3601.895 | 1231.609 |
| Zero-shot LST | 747.978 | 437.010 | 755.625 | 442.511 |

Table II shows the performance with the CT dataset. One-shot CoT is found to achieve the best performance, while one-shot SARIMA performs the worst. This contrasting result may imply two important aspects in time series forecasting with LLMs. A well-designed example works as a good reference to lead the LLM to complete the task better. On the other hand, a one-shot example with a certain type of prompting may degrade performance. SARIMA prompting consists of quite detailed procedures. Thus, when a detailed demonstration is not given in a way that the LLM can understand accurately, the LLM may have difficulty in understanding the procedures in the prompt and demonstration example. This can be particularly problematic when all the details of each procedure are not given in detail. It is expected that reasoning with CoT gives the opportunity to take world knowledge into account, which can make predictions more consistent with physical phenomena. When the responses were manually analyzed, there were few cases where one-shot CoT explicitly exploited its internal knowledge, while it often kept the procedure for forecasting as data inspection, calculation of key statistics, analysis of trends, and prediction of the next value. One illustrative example of exploiting world knowledge can be given as: "Therefore, a reasonable estimate for October 1, 2019, could be around the average of the recent days, but with considerations for fluctuation, we might predict slightly higher due to typical October weather patterns," where the typical October weather pattern was considered. Similar characteristics can be observed in Table III for the ECL dataset.

TABLE IV. RMSE PERFORMANCE WITH IHEPC DATASET

| Forecasting step | 1 | 2 | 3 | 4 | 5 | 6 |
|---|---|---|---|---|---|---|
| Baseline | 4.249 | **4.431** | 4.765 | **4.625** | **4.769** | **4.701** |
| Zero-shot CoT | 4.300 | 4.477 | **4.757** | 4.770 | 4.848 | 4.753 |
| One-shot CoT | 4.285 | 4.672 | 5.013 | 5.102 | 5.253 | 5.419 |
| Zero-shot PaS+ | 4.673 | 4.779 | 4.997 | 5.016 | 5.137 | 5.052 |
| Zero-shot SARIMA | 4.388 | 4.716 | 5.025 | 4.992 | 5.159 | 5.115 |
| One-shot SARIMA | 6.534 | 6.850 | 7.117 | 7.324 | 7.691 | 8.254 |
| Zero-shot LST | **3.809** | 4.514 | 4.929 | 4.966 | 5.701 | 6.226 |

TABLE V. MAE PERFORMANCE WITH IHEPC DATASET

| Forecasting step | 1 | 2 | 3 | 4 | 5 | 6 |
|---|---|---|---|---|---|---|
| Baseline | 3.112 | 3.303 | 3.604 | **3.505** | **3.633** | 3.565 |
| Zero-shot CoT | 3.167 | 3.381 | 3.665 | 3.659 | 3.742 | 3.622 |
| One-shot CoT | 3.116 | 3.531 | 3.848 | 3.860 | 3.962 | 4.147 |
| Zero-shot PaS+ | 3.160 | 3.303 | **3.561** | 3.523 | 3.641 | **3.563** |
| Zero-shot SARIMA | 3.245 | 3.553 | 3.872 | 3.848 | 3.935 | 3.888 |
| One-shot SARIMA | 4.479 | 4.739 | 5.017 | 5.048 | 5.267 | 5.433 |
| Zero-shot LST | **2.696** | **3.229** | 3.661 | 3.623 | 3.942 | 3.841 |

Due to the variation in responses, it was extremely difficult to extract the prediction results without missing data. It might have been easier to retrieve the predicted value when a strong requirement on the response format was imposed on the prompt. However, to avoid possible interference with forecasting itself, rather weak requests on the format, such as "please answer the predicted value only" or "show the Final Answer with predicted value only," were imposed. Many different regular expressions were tried to be tailored to a specific dataset and prompting. Nonetheless, there were cases where a significant portion of the response could not be considered for performance evaluation, as shown in [18]. Depending on dataset and prompting, the missing rates were significantly different. To avoid misleading results from differences in the composition of the data for evaluation, forecasting performance was evaluated for the data samples where predictions were successfully retrieved for all promptings, with RMSE* and MAE* corresponding to those results. It can be observed that due to the large number of data samples for each dataset, there is no significant difference depending on the composition of data samples for performance evaluation.

Multi-step forecasting with long time series was evaluated with the IHEPC dataset, where each series corresponds to 96 consecutive hourly averaged current usage. Due to the nature of electrical usage in a house, relatively strong seasonality can be expected with this dataset. For one-shot CoT and one-shot SARIMA, the same examples designed for the short time series were used to maintain consistency in evaluation. RMSE and MAE with different promptings over 6 steps are shown in Table IV and V respectively. Interestingly, LST prompting achieved the best performance for steps 1, and 2 in MAE performance, while baseline prompting achieved the best performance for steps 2, 4, 5, and 6 in RMSE. It is conjectured that short-term predictions based on internal knowledge of forecasting may be effective, while long-term predictions may not be effective enough to compensate for the prediction error with increasing steps. It is also observed that many responses with LST prompting have separate predictions for short-term and long-term without being integrated into the final predictions which results in a large missing rate. It is noted that PaS+ prompting achieves near-best MAE performance, while RMSE performance is relatively worse. PaS+ prompting may have a relatively larger number of significant errors compared to other prompting methods."

**Qualitative Analysis**: When all quantitative results are considered, it appears that there is no single powerful prompting method that provides good forecasting performance for all datasets. Therefore, it is meaningful to summarize the issues in the responses. Detailed analysis can be found online [18].

1. Quantized forecasting was observed with the ECL dataset.

2. The response from one-shot SARIMA often fails to subtract the trend component when dealing with seasonality, resulting in a significant offset in the final prediction. To accurately predict the seasonality component, the trend component should be subtracted from the series.

3. Regardless of prompting, when an LLM performs algebraic operations with a significant number of operands, it often fails to calculate the operation correctly.

4. Many promptings often fail to follow its planned reasoning steps, and instead rely only on recent trends, leading to inconsistencies between the prompt's direction and the actual calculation.

5. Many promptings with the IHEPC dataset often fail to extract meaningful information associated with long-term variation over the whole sequence.

6. LST prompting sometimes fails to generate the next 6 predicted values. It predicts 5 steps only

7. Forecasting often depends on recent values only.

## V. Conclusions

In this paper, we characterized the effects of several different prompting methods for time series forecasting. Although not all existing promptings were considered, some frequently observed patterns in the responses associated with time series forecasting were identified. Reasoning steps involving simple algebraic operations often fail to calculate accurately, which is conjectured to be due to interference from the previously generated contexts. Moreover, if a reasoning step involves a parametric model with parameters determined from inaccurate calculations, it results in large errors. GPT-4o-mini is also observed to have a semantic gap between domain-specific and general terminology. For example, trends are often treated as short-term characteristics rather than long-term ones. Regardless of the prompting method, seasonality is rarely exploited properly to reduce forecasting errors effectively. In many cases, forecasting is based on several recent values only.

Prompting that requires a certain structure for reasoning appears to have internal conflicts between the reasoning steps described in the prompt and its own reasoning steps learned from training, while simple prompting with key contextual information often provides reliable responses. Additionally, prompting with detailed procedures and requirements often leads to misunderstandings of the task associated with subtasks or integrating the subtask results. Considering the cost and potential degradation with inappropriate prompting, it is expected that providing proper context information on time series can be more important than asking for a specific type of reasoning, while there is potential that a specifically crafted prompting can effectively lead to better responses.

Several interesting aspects need further attention. LLMs may have inherently limited performance in time series forecasting since they are simply trained to generate the next token based on preceding tokens. This can be especially the case when tokenization cannot effectively represent the relationships among numeric values in time series. Thus, proper fine-tuning is required to fit time series forecasting, which requires a training corpus that addresses the shortcomings of LLMs for time series forecasting. Moreover, joint fine-tuning and prompting may have potential for further improvement, such that fine-tuning can support the requirements of the prompting better, and datasets for fine-tuning can be more fitted to a specific type of prompting. Additionally, fundamental limitations by tokenization need to be addressed further. Proper scaling and offsetting can be an alternative way to deal with this issue approximately, rather than having an adaptor module for patching, which incurs additional complexity for training."